\PassOptionsToPackage{hypertexnames=false}{hyperref}
\documentclass[pdflatex,sn-mathphys-num]{sn-jnl}

\usepackage{amsmath,amssymb}
\usepackage{booktabs}
\usepackage{graphicx}
\usepackage{microtype}
\usepackage{algorithm}
\usepackage{algorithmicx}
\usepackage{algpseudocode}
\usepackage{tikz}
\usetikzlibrary{arrows.meta,positioning}

\begin{document}

\title[Difficulty-Aware Sample Allocation]{Difficulty-Aware Sample Allocation for Adaptive Data Augmentation in Semantic Segmentation}

\author*[1]{\fnm{Olasimbo Ayodeji} \sur{Arigbabu}}\email{oa.arigbabu@gmail.com}

\author[2]{\fnm{Abimbola Ismail} \sur{Arigbabu}}\email{abimbola.arigbabu@oouagoiwoye.edu.ng}

\affil*[1]{\orgname{Independent Researcher}, \country{Germany}}

\affil[2]{\orgdiv{Department of Science and Technology Education}, \orgname{Olabisi Onabanjo University}, \orgaddress{\city{Ago-Iwoye}, \state{Ogun State}, \country{Nigeria}}}

\abstract{Data augmentation is a standard component of modern semantic segmentation pipelines, but most augmentation techniques allocate transformations uniformly across training samples or adapt to a single difficulty signal such as loss. This ignores the fact that segmentation difficulty is multi-factorial, since ambiguous predictions, persistent optimization errors, rare classes, and complex object boundaries can each make a sample informative in different ways. This paper introduces Difficulty-Aware Sample Allocation (DASA), an architecture-agnostic framework that assigns stronger augmentation to samples estimated to be more difficult. DASA combines prediction ambiguity, training loss, class rarity, and boundary complexity into a normalized difficulty score, then maps that score to sample-specific augmentation strength during iterative training. Experiments on Oxford-IIIT Pet and binary Pascal VOC segmentation with U-Net, DeepLabV3, and SegFormer-B0 show that DASA improves over standard training and is competitive with or stronger than single-signal adaptive baselines. On Oxford-IIIT Pet, DASA improves DeepLabV3 from 0.633 to 0.740 mIoU. On binary Pascal VOC, DASA obtains the best foreground IoU for all three evaluated architectures. These results attest to the value of multi-factor difficulty estimation as a practical mechanism for directing augmentation where it is most useful.}


\maketitle

\section{Introduction}

Semantic segmentation models have achieved strong performance across medical imaging, autonomous driving, robotics, remote sensing, and general visual recognition. Unlike image classification, segmentation requires a model to assign a semantic label to every pixel. This makes the task sensitive not only to object presence, but also to boundary precision, local texture, object scale, occlusion, and the spatial relationship between foreground and background. Architectures such as U-Net~\citep{ronneberger2015unet}, DeepLabV3~\citep{chen2017rethinking}, and transformer-based models such as SegFormer~\citep{xie2021segformer} demonstrate that dense prediction can benefit from both convolutional and attention-based representations. Despite these advances, segmentation performance remains highly dependent on training data diversity and on the model's ability to generalize to changes in appearance, scale, pose, and shape.

Data augmentation is one of the most widely used tools for improving generalization~\citep{shorten2019survey,lewy2023overview}. Geometric transformations, photometric perturbations, random crops, erasing strategies, and mixing-based methods expose the model to a broader effective training distribution~\citep{zhang2018mixup,yun2019cutmix,cubuk2020randaugment}. For segmentation, augmentation must also preserve the alignment between an image and its label map. Segmentation-specific augmentation methods such as ClassMix and Copy-Paste reflect this need to modify images while respecting semantic or instance structure~\citep{olsson2020classmix,ghiasi2020copypaste}. A transformation that is useful for classification may be harmful for segmentation if it weakens boundary correspondence, removes small structures, or produces unrealistic object geometry. Thus, augmentation in segmentation is not only a question of increasing data variation. It is also a question of deciding how much perturbation a labeled sample can usefully support.

In most pipelines, augmentation is applied uniformly or according to a fixed schedule. Automated and simplified policy methods such as AutoAugment and RandAugment improve policy selection, but the resulting policies are still typically applied at the dataset level~\citep{cubuk2019autoaugment,cubuk2020randaugment}. Once the augmentation policy is chosen, every image is exposed to approximately the same distribution of transformations. This design is simple and often effective, but it hides an important assumption. It assumes that all samples benefit equally from the same augmentation budget. In practice, segmentation datasets are rarely homogeneous. Images differ in object size, pose, class frequency, annotation structure, boundary density, and current model uncertainty. Treating these images identically can make augmentation inefficient, and this assumption is often weak for segmentation.

Some images contain large, centered objects with simple boundaries; others contain small objects, rare classes, occlusions, clutter, fine structures, or highly irregular contours. A uniform augmentation policy can waste effort on easy examples while under-serving samples that are more informative for model improvement. Existing adaptive strategies, including hard example mining and curriculum learning, partially address this issue by emphasizing difficult samples~\citep{bengio2009curriculum,shrivastava2016ohem}. Many such methods, however, rely on a single difficulty indicator such as loss or confidence, whereas segmentation difficulty is richer than any single signal.

The practical consequence is that augmentation should be treated as a resource to be assigned, not only as a fixed preprocessing step. This is especially important in dense prediction, where the usefulness of a sample may depend on model uncertainty, persistent pixel-wise errors, semantic imbalance, and shape structure at the same time. A sample may be easy according to loss but still contain a rare class. Another sample may contain common labels but remain difficult because its boundary occupies many fine structures. These cases motivate an allocation mechanism that combines complementary evidence rather than selecting one signal as sufficient.

This perspective differs from choosing a stronger global policy. Prior augmentation-policy methods show that stronger or better selected transformations can improve generalization and robustness~\citep{cubuk2019autoaugment,hendrycks2019augmix,muller2021trivialaugment}. However, a global increase in augmentation can also degrade performance when easy or already well-represented samples are over-perturbed. Conversely, weak augmentation can preserve label fidelity but may fail to challenge samples that contain rare objects, uncertain predictions, or complex contours. The central problem addressed in this paper is therefore sample allocation. Given a fixed augmentation family, the method should decide which samples should receive stronger transformations and which samples should remain closer to their original form.

This paper proposes Difficulty-Aware Sample Allocation (DASA), a simple adaptive augmentation framework for semantic segmentation. DASA estimates sample difficulty using four complementary signals, namely prediction ambiguity, optimization difficulty, class rarity, and boundary complexity. The resulting difficulty score controls the strength of augmentation applied to each sample during training. The approach does not require changes to the segmentation architecture and can be used with convolutional or transformer-based models.

The main contributions are as follows.
\begin{itemize}
    \item A multi-factor difficulty estimator for semantic segmentation that combines ambiguity, loss, rarity, and boundary structure.
    \item A sample-level augmentation allocation rule that maps normalized difficulty to augmentation strength.
    \item An architecture-agnostic implementation evaluated with U-Net, DeepLabV3, and SegFormer-B0.
    \item A controlled empirical study on Oxford-IIIT Pet and binary Pascal VOC with single-factor ablations that separate the effect of loss, rarity, and boundary information.
\end{itemize}

\section{Related Work}

Data augmentation is widely recognized as a core mechanism for improving generalization in deep learning. Surveys such as Shorten and Khoshgoftaar~\citep{shorten2019survey} and Lewy and Mandziuk~\citep{lewy2023overview} show that the literature has developed a broad vocabulary of transformations, including geometric operations, color-space perturbations, kernel filters, random erasing, image mixing, generative augmentation, and automated policy search. This range is valuable, but it also exposes a limitation in many practical pipelines, since augmentation is often treated as a dataset-level processing. Once the policy is chosen, the same distribution of transformations is usually applied across all samples, even though segmentation datasets contain highly heterogeneous examples.

Mixing-based augmentation shows this limitation clearly. MixUp~\citep{zhang2018mixup} and CutMix~\citep{yun2019cutmix} improve regularization by constructing synthetic examples from multiple images, and segmentation-specific variants such as ClassMix~\citep{olsson2020classmix} modify this idea to respect predicted semantic regions. Copy-Paste augmentation similarly changes the composition of training images by inserting object instances into new contexts~\citep{ghiasi2020copypaste}. These methods are powerful because they expand the training distribution, but they primarily answer the question of \emph{what} augmented example to create. They do not directly answer \emph{which} samples should receive more augmentation effort. DASA targets this second question by estimating the difficulty of each sample and using that estimate to allocate augmentation intensity.

Automated augmentation methods address a related but distinct problem. AutoAugment~\citep{cubuk2019autoaugment} searches for effective augmentation policies, while Fast AutoAugment~\citep{lim2019fastautoaugment}, Population Based Augmentation~\citep{ho2019pba}, RandAugment~\citep{cubuk2020randaugment}, TrivialAugment~\citep{muller2021trivialaugment}, and AugMix~\citep{hendrycks2019augmix} reduce search cost, simplify the policy space, or improve robustness. The strength of this line of work is that it reduces manual policy design, meanwhile, its limitation for the present problem is that the learned or sampled policy is still generally global, regularizing the dataset as a whole rather than allocating different augmentation budgets to different samples. Recent segmentation studies reinforce the importance of augmentation design. Schwonberg et al.~\citep{schwonberg2023augmentationdg} show that combinations of simple rule-based augmentations can be competitive for domain generalization in semantic segmentation; Che et al.~\citep{che2025enhancedgenerative} explore controllable diffusion-based synthetic augmentation while preserving segmentation structure; and Ran et al.~\citep{ran2025adaptivespatial} adapt spatial augmentation in semi-supervised segmentation using entropy. These works support the view that augmentation should be controlled carefully, but DASA differs by combining multiple difficulty signals to drive sample-level allocation.

The idea that training examples should not be treated uniformly also appears in curriculum learning and hard-example mining. Curriculum learning orders examples to improve optimization, often progressing from easy to difficult samples~\citep{bengio2009curriculum}. Self-paced learning makes the curriculum adaptive by allowing the model to select examples according to its current competence~\citep{kumar2010selfpaced}. Online batch selection and Online Hard Example Mining prioritize examples with large losses or high training difficulty~\citep{loshchilov2015online,shrivastava2016ohem}. These methods establish that sample difficulty is useful, but most of them use difficulty to alter example ordering, sampling, or loss contribution. DASA instead uses difficulty to control augmentation strength. This distinction is important because a difficult segmentation sample may benefit not only from being seen more often, but from being seen under more diverse transformations.

Segmentation adds further reasons to move beyond a single difficulty signal. Class imbalance motivates class-aware rebalancing methods such as class-balanced loss~\citep{cui2019classbalanced}. Boundary-aware losses show that contour structure contains information not captured by region-level objectives alone~\citep{kervadec2019boundary}. Uncertainty and loss capture model-dependent difficulty, while rarity and boundary complexity capture data-dependent semantic and structural difficulty. Existing methods usually emphasize one of these factors at a time. DASA is motivated by the gap between these separate views, since segmentation difficulty is multi-factorial and augmentation allocation should account for ambiguity, optimization difficulty, class rarity, and boundary complexity jointly.

\section{Method}

\subsection{Problem Setup}

Let $\mathcal{D}=\{(x_i,y_i)\}_{i=1}^{N}$ be a semantic segmentation dataset, where $x_i$ is an image and $y_i$ is a pixel-level label map. A conventional augmentation pipeline applies transformations sampled from a policy $\mathcal{T}$ with fixed probabilities. DASA instead assigns each sample a scalar augmentation strength
\begin{equation}
    s_i \in [0,1],
\end{equation}
where larger values correspond to stronger or more frequent transformations. The strength $s_i$ is derived from a multi-factor difficulty score.

The interval $[0,1]$ makes the allocation independent of the specific augmentation library. A value near zero means that the sample receives only weak transformations, such as a low-probability flip or mild color jitter. A value near one means that the same sample is exposed to stronger transformations, such as larger rotations, stronger color changes, or a higher probability of applying multiple operations. Thus, $s_i$ should be interpreted as an augmentation budget assigned to sample $i$, not as a model prediction.

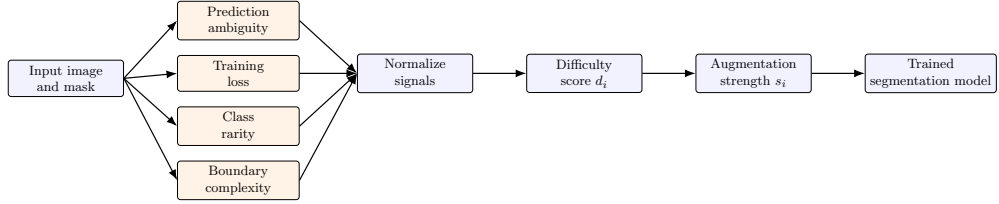
\begin{figure}[ht]
    \centering
    \resizebox{\linewidth}{!}{%
    \begin{tikzpicture}[
        font=\small,
        node distance=0.95cm and 1.25cm,
        process/.style={
            draw,
            rounded corners=2pt,
            align=center,
            minimum width=2.7cm,
            minimum height=0.85cm,
            fill=blue!5
        },
        signal/.style={
            draw,
            rounded corners=2pt,
            align=center,
            minimum width=2.85cm,
            minimum height=0.65cm,
            fill=orange!10
        },
        arrow/.style={-{Latex[length=2.5mm]}, thick}
    ]
        \node[process] (input) {Input image\\and mask};

        \node[signal, right=of input, yshift=1.35cm] (ambiguity) {Prediction\\ambiguity};
        \node[signal, below=0.35cm of ambiguity] (loss) {Training\\loss};
        \node[signal, below=0.35cm of loss] (rarity) {Class\\rarity};
        \node[signal, below=0.35cm of rarity] (boundary) {Boundary\\complexity};

        \node[process, right=1.35cm of loss] (normalize) {Normalize\\signals};
        \node[process, right=of normalize] (score) {Difficulty\\score $d_i$};
        \node[process, right=of score] (strength) {Augmentation\\strength $s_i$};
        \node[process, right=of strength] (model) {Trained\\segmentation model};

        \draw[arrow] (input.east) -- (ambiguity.west);
        \draw[arrow] (input.east) -- (loss.west);
        \draw[arrow] (input.east) -- (rarity.west);
        \draw[arrow] (input.east) -- (boundary.west);

        \draw[arrow] (ambiguity.east) -- (normalize.west);
        \draw[arrow] (loss.east) -- (normalize.west);
        \draw[arrow] (rarity.east) -- (normalize.west);
        \draw[arrow] (boundary.east) -- (normalize.west);

        \draw[arrow] (normalize) -- (score);
        \draw[arrow] (score) -- (strength);
        \draw[arrow] (strength) -- (model);
    \end{tikzpicture}%
    }
    \caption{DASA pipeline. Each training sample is evaluated using prediction ambiguity, training loss, class rarity, and boundary complexity. The normalized signals are combined into a difficulty score, which determines the augmentation strength used during model training.}
    \label{fig:dasa_pipeline}
\end{figure}

DASA does not replace the underlying augmentation policy. Instead, it controls how strongly that policy is applied to each sample. This design keeps the framework compatible with common segmentation pipelines. The same base transformations can be used for every method, while only the sample-level strength changes. As a result, performance differences can be attributed to the allocation rule rather than to a completely different set of transformations. Figure~\ref{fig:dasa_pipeline} summarizes the overall DASA workflow from input sample to difficulty estimation, augmentation-strength assignment, and model training.

\subsection{Difficulty Signals}

\subsubsection{Prediction ambiguity}
This signal captures model uncertainty. Given $K$ stochastic forward passes, for example using dropout at inference time, let $p_i^{(k)}$ denote the predicted class probability map for sample $i$. The mean prediction is
\begin{equation}
    \bar{p}_i = \frac{1}{K}\sum_{k=1}^{K}p_i^{(k)}.
\end{equation}
DASA averages the $K$ probability maps to estimate the model's typical belief about each pixel. If repeated passes produce similar probability maps, then $\bar{p}_i$ will be concentrated on one class at most pixels. If the model is unstable, the averaged probability vector will be more spread out across classes.
DASA computes ambiguity with normalized predictive entropy.
\begin{equation}
    A_i = \frac{1}{|\Omega_i|\log C}\sum_{u \in \Omega_i}
    H\left(\bar{p}_i(u)\right),
\end{equation}
where $\Omega_i$ is the pixel grid, $C$ is the number of classes, and $H(\cdot)$ is categorical entropy.
The summation averages entropy over all pixels in the image. Dividing by $\log C$ normalizes the entropy so that $A_i$ is comparable across datasets with different numbers of classes. Low entropy means the model assigns high probability to one class, so the prediction is confident. High entropy means the model distributes probability across multiple classes, so the sample is ambiguous and receives a larger difficulty contribution.

\subsubsection{Optimization difficulty}
This signal is measured using per-sample segmentation loss.
\begin{equation}
    L_i = \frac{1}{|\Omega_i|}\sum_{u \in \Omega_i}
    \ell\left(f_\theta(x_i)_u, y_i(u)\right),
\end{equation}
where $f_\theta$ is the segmentation model and $\ell$ is categorical cross-entropy loss~\citep{goodfellow2016deep}.
This term measures how poorly the current model fits the labeled mask. The prediction $f_\theta(x_i)_u$ is the class distribution predicted at pixel $u$, and $y_i(u)$ is the corresponding ground-truth label. Averaging over pixels prevents larger images or masks with more labeled pixels from automatically receiving larger scores. A high $L_i$ indicates that the model is still making many confident or repeated mistakes on sample $i$, so additional augmented views may be useful.

\subsubsection{Class rarity}
This signal increases the difficulty score for samples containing underrepresented labels. If $q_c$ is the empirical pixel frequency of class $c$, the rarity score is
\begin{equation}
    R_i = \frac{1}{|\mathcal{C}_i|}\sum_{c \in \mathcal{C}_i}
    \frac{1}{q_c+\epsilon},
\end{equation}
where $\mathcal{C}_i$ is the set of classes present in $y_i$ and $\epsilon$ prevents division by zero.
The inverse-frequency term gives larger values to classes that occupy fewer pixels in the training set. Averaging over $\mathcal{C}_i$ produces one rarity value per image rather than one value per class. Consequently, an image containing a rare class receives a larger $R_i$ even if the model's current loss is not unusually high. This is useful because rare classes can be under-trained even when their individual examples are not the hardest examples by loss.

\subsubsection{Boundary complexity}
This signal measures structural difficulty in the target mask. Let $\nabla y_i$ be a binary edge map extracted from the segmentation mask. DASA estimates boundary complexity as contour density.
\begin{equation}
    B_i = \frac{1}{|\Omega_i|}\sum_{u \in \Omega_i}\mathbb{1}\left[\nabla y_i(u)>0\right].
\end{equation}
Samples with more irregular or dense contours receive larger values.
Here, $\mathbb{1}[\cdot]$ denotes the indicator function, which contributes one whenever pixel $u$ lies on a mask boundary and zero otherwise. Dividing by the number of pixels converts the boundary count into a normalized boundary density. This makes $B_i$ larger for masks with fine structures, thin parts, holes, or highly non-convex shapes. Such samples are often difficult in segmentation because small spatial errors can sharply reduce IoU near object boundaries.

\subsection{Difficulty Aggregation}
Each signal is min-max normalized over the training set or over the current difficulty-estimation pass.
\begin{equation}
    \tilde{Z}_i = \frac{Z_i - \min_j Z_j}{\max_j Z_j - \min_j Z_j + \epsilon},
    \qquad Z \in \{A,L,R,B\}.
\end{equation}
This step places ambiguity, loss, rarity, and boundary complexity on the same numerical scale before they are combined. Without normalization, a signal with a naturally larger numeric range could dominate the score even if it is not more important. The final difficulty score is then computed as a convex combination.
\begin{equation}
    d_i = \alpha \tilde{A}_i + \beta \tilde{L}_i + \gamma \tilde{R}_i + \delta \tilde{B}_i,
    \qquad
    \alpha+\beta+\gamma+\delta=1.
\end{equation}
In the experiments, the weights are set to
\begin{equation}
    \alpha=0.35,\quad \beta=0.35,\quad \gamma=0.20,\quad \delta=0.10.
\end{equation}
These values place the greatest emphasis on uncertainty and optimization difficulty while retaining semantic and structural information.
Because the weights sum to one and all four inputs are normalized, $d_i$ remains interpretable as a relative difficulty score. A sample receives a high score only when one or more normalized signals are high. For example, an image with uncertain predictions and high loss will receive a large score even if its classes are common, while an image containing a rare class can still receive additional augmentation even if its current loss is moderate.

\subsection{Design Rationale}

The four difficulty signals are selected to cover distinct sources of segmentation difficulty. Ambiguity and loss are model-dependent. They change as the current network learns and therefore capture the state of training. Rarity and boundary complexity are data-dependent. They capture properties of the labeled mask that may remain important even when the current loss is not large. Combining both types of evidence reduces the risk that augmentation is driven only by temporary optimization noise.

The linear aggregation is intentionally simple. It makes the contribution of each signal visible and supports direct ablation against loss-only, rarity-only, and boundary-only variants. More complex combinations, such as learned weighting networks, could increase flexibility but would also make it harder to determine whether the gain comes from difficulty-aware allocation or from additional model capacity. For this reason, the present study treats DASA as a transparent allocation heuristic and evaluates its usefulness empirically.

\subsection{Augmentation Allocation}
The normalized difficulty score is mapped to an augmentation probability or strength.
\begin{equation}
    s_i = s_{\min} + d_i(s_{\max}-s_{\min}).
\end{equation}
When sample $i$ is drawn during training, its augmentation parameters are sampled according to $s_i$. In practice, larger values increase the probability or magnitude of geometric and photometric transformations. Easy samples therefore remain lightly augmented, while difficult samples receive more aggressive augmentation.
This equation linearly interpolates between the weakest and strongest allowed augmentation settings. If $d_i=0$, then $s_i=s_{\min}$ and the sample receives the minimum augmentation budget. If $d_i=1$, then $s_i=s_{\max}$ and the sample receives the maximum budget. Intermediate values produce proportional augmentation strength. The constants $s_{\min}$ and $s_{\max}$ prevent the method from completely removing augmentation from easy samples or applying unrealistically severe augmentation to hard samples.

\begin{figure}[ht]
    \centering
    \includegraphics[width=\linewidth]{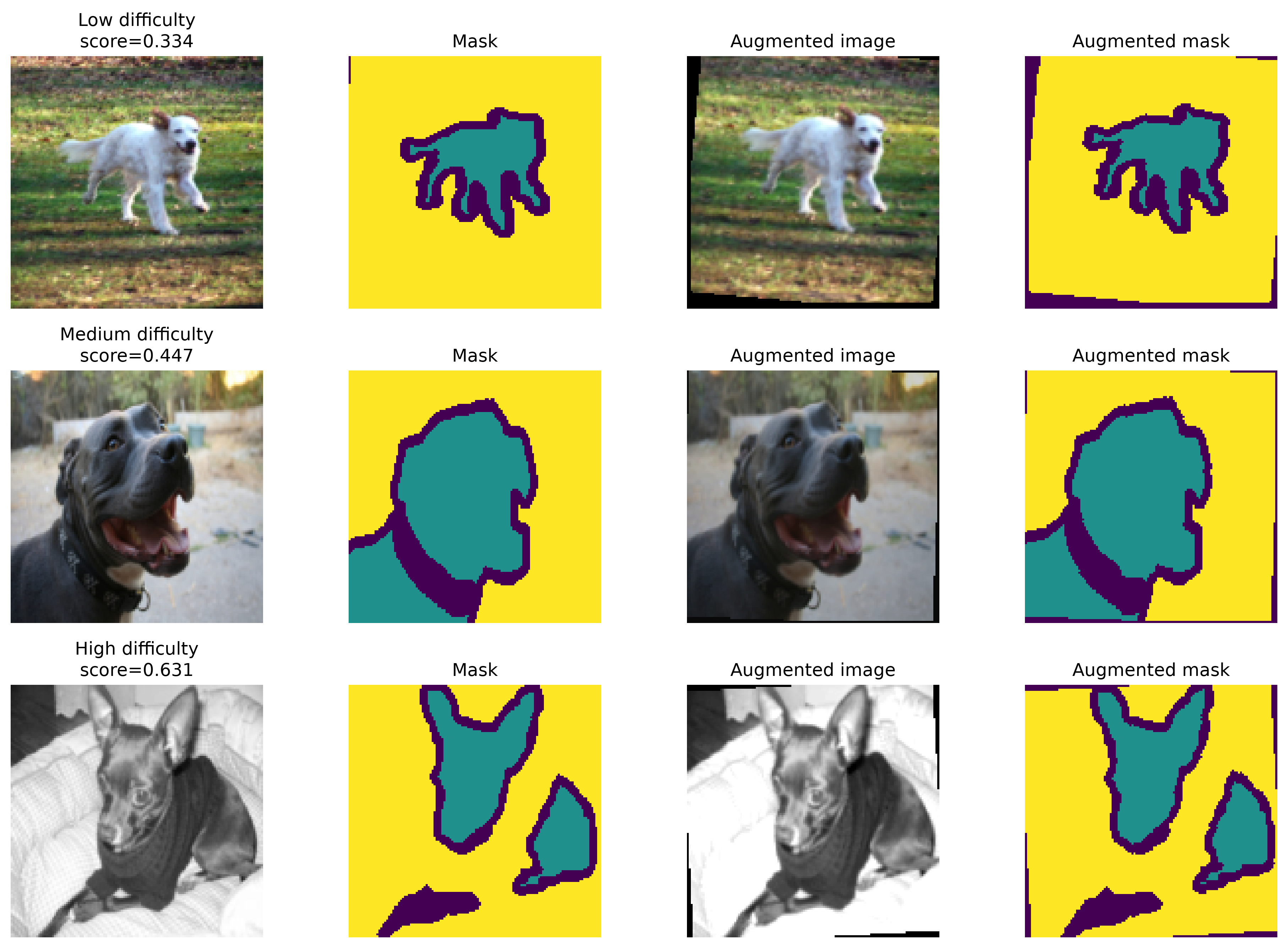}
    \caption{Examples of difficulty-aware augmentation on Oxford-IIIT Pet using DeepLabV3. Low-, medium-, and high-difficulty samples receive different augmentation strengths while preserving image-mask correspondence.}
    \label{fig:difficulty_examples_pet}
\end{figure}

Figure~\ref{fig:difficulty_examples_pet} illustrates the allocation behavior on Oxford-IIIT Pet samples. As the difficulty score increases, the corresponding augmented view is subjected to stronger perturbation while the image and mask remain spatially aligned.

\subsection{Training Procedure}

DASA alternates between model training and difficulty estimation. At the beginning of each round, the current model evaluates the training samples to update the four difficulty signals. These signals define the next round's augmentation strengths. The model is then trained using the updated allocation. Algorithm~\ref{alg:dasa} summarizes the complete procedure.

\begin{algorithm}[t]
\caption{Difficulty-Aware Sample Allocation (DASA)}
\label{alg:dasa}
\begin{algorithmic}[1]
\Require Training set $\mathcal{D}=\{(x_i,y_i)\}_{i=1}^{N}$; segmentation model $f_\theta$; augmentation policy $\mathcal{T}$; number of rounds $M$; stochastic passes $K$; weights $\alpha,\beta,\gamma,\delta$; strength bounds $s_{\min},s_{\max}$
\Ensure Trained model parameters $\theta$
\State Initialize model parameters $\theta$
\State Initialize sample strengths $s_i \gets s_{\min}$ for all $i \in \{1,\ldots,N\}$
\For{$m = 1$ to $M$}
    \State Train $f_\theta$ for one round using augmentations sampled from $\mathcal{T}$ with sample strengths $\{s_i\}_{i=1}^{N}$
    \For{each training sample $(x_i,y_i) \in \mathcal{D}$}
        \State Run $K$ stochastic forward passes and compute $\bar{p}_i = \frac{1}{K}\sum_{k=1}^{K}p_i^{(k)}$
        \State Compute ambiguity $A_i$ from normalized predictive entropy
        \State Compute optimization difficulty $L_i$ from per-sample segmentation loss
        \State Compute class rarity $R_i$ from inverse class frequencies in $y_i$
        \State Compute boundary complexity $B_i$ from mask contour density
    \EndFor
    \State Normalize each signal to obtain $\tilde{A}_i,\tilde{L}_i,\tilde{R}_i,\tilde{B}_i$ for all samples
    \For{each training sample $(x_i,y_i) \in \mathcal{D}$}
        \State Compute difficulty $d_i \gets \alpha \tilde{A}_i + \beta \tilde{L}_i + \gamma \tilde{R}_i + \delta \tilde{B}_i$
        \State Update augmentation strength $s_i \gets s_{\min} + d_i(s_{\max}-s_{\min})$
    \EndFor
\EndFor
\State \Return $\theta$
\end{algorithmic}
\end{algorithm}

\section{Experimental Setup}

\subsection{Datasets}

\subsubsection{Oxford-IIIT Pet}
The Oxford-IIIT Pet dataset contains pet images with pixel-level trimap annotations~\citep{parkhi2012cats}. The trimap setting includes foreground animal regions, object boundary regions, and background. This dataset is useful for evaluating DASA because many images contain fine fur boundaries, pose variation, and differences in object scale. The experiments evaluate three-class segmentation using mean Intersection over Union (mIoU), mean Dice (mDice), class-wise IoU, and runtime.

\subsubsection{Pascal VOC Binary}
Pascal VOC segmentation labels are converted to a binary foreground-background task~\citep{everingham2010pascal}. All annotated object categories are treated as foreground and the remaining pixels are treated as background. This setting emphasizes object localization and foreground recovery. It also creates a strong imbalance between background and foreground pixels, making it a useful test case for difficulty allocation.

\subsection{Evaluation Metrics}

The primary metric is mIoU, which measures the overlap between predicted and ground-truth regions and is standard in semantic segmentation. Mean Dice is also reported because it is sensitive to foreground recovery and is commonly used when class imbalance is present. For Oxford-IIIT Pet, class-wise IoU and Dice are used to inspect the trimap classes. For binary Pascal VOC, foreground IoU is emphasized because background performance can be high even when the model misses important object pixels.

\subsection{Models and Baselines}

Experiments use U-Net, DeepLabV3, and SegFormer-B0. The following training strategies are compared.
\begin{itemize}
    \item \textbf{Baseline} standard training without adaptive allocation.
    \item \textbf{Strong uniform} stronger augmentation applied uniformly, following the common practice of increasing augmentation magnitude globally rather than per sample~\citep{cubuk2020randaugment}.
    \item \textbf{Random weighted} random sample-specific augmentation weights, used as a control to test whether any non-uniform allocation helps even without difficulty information.
    \item \textbf{Loss-only} allocation based only on optimization difficulty, motivated by hard-example mining methods that prioritize high-loss or difficult examples~\citep{shrivastava2016ohem}.
    \item \textbf{Rarity-only} allocation based only on class rarity, motivated by class-rebalancing methods for long-tailed recognition~\citep{cui2019classbalanced}.
    \item \textbf{Boundary-only} allocation based only on boundary complexity, motivated by boundary-aware segmentation objectives that emphasize contour structure~\citep{kervadec2019boundary}.
    \item \textbf{DASA} the proposed multi-factor allocation method.
\end{itemize}
The loss-only, rarity-only, and boundary-only strategies are single-factor ablations inspired by these research directions; they are not intended as exact reproductions of the cited methods.

The three architectures cover different model families and capacities. U-Net represents a compact encoder-decoder convolutional model. DeepLabV3 represents a stronger convolutional segmentation architecture with atrous spatial context modeling. SegFormer-B0 represents a lightweight transformer-based segmentation model. Evaluating all three helps test whether DASA depends on a specific architecture or remains useful across different representation types.

\subsection{Experimental Protocol}

All methods are compared within the same dataset and architecture setting. The baseline uses the standard training pipeline without adaptive allocation. Strong uniform uses a globally stronger augmentation setting for every sample. Random weighted keeps the allocation non-uniform but removes difficulty information. The single-factor variants isolate the contribution of one signal at a time. DASA uses the combined score from ambiguity, loss, rarity, and boundary complexity.

For adaptive methods, training proceeds in rounds. At the start of each round, the current model estimates sample difficulty on the training set. The resulting scores define the augmentation strength used in the next training round. This round-based design avoids updating augmentation strength after every mini-batch, which would be more expensive and less stable. Runtime is reported to make the cost of adaptive allocation visible alongside accuracy.

All experiments are conducted using image size of $128 \times 128$. Baseline models are trained for 20 epochs and adaptive methods use 3 rounds with 3 training epochs per round. The learning rate is 0.001, weight decay is 0.001, and DASA uses 4 Monte Carlo passes for ambiguity estimation. The batch size is 20 for U-Net and DeepLabV3, while SegFormer-B0 uses batch size 8 because of its higher memory demand. 

\section{Results}

\subsection{Oxford-IIIT Pet}

Table~\ref{tab:pet_results} reports Oxford-IIIT Pet performance. DASA improves all three architectures over their non-adaptive baselines and obtains the best mIoU for U-Net, DeepLabV3, and SegFormer-B0. The largest gain occurs for DeepLabV3, where mIoU increases from 0.633 to 0.740. For SegFormer-B0, DASA is slightly stronger than the loss-only ablation, indicating that the combined difficulty estimate can improve on the strongest single-signal strategy.

The Pet results also show that stronger augmentation is not automatically beneficial. Strong uniform reduces performance for U-Net and SegFormer-B0, even though it improves DeepLabV3. This supports the central motivation of the paper. Augmentation strength should not only be increased globally, because the same intensity can help one architecture or sample distribution while hurting another. DASA avoids this all-or-nothing behavior by reserving stronger transformations for samples estimated to be more informative.

\begin{table}[t]
\centering
\caption{Oxford-IIIT Pet results. Best mIoU per model is bolded.}
\label{tab:pet_results}
\begin{tabular}{llccc}
\toprule
Model & Method & mIoU & mDice & Runtime (min) \\
\midrule
U-Net & Baseline & 0.692 & 0.794 & 5.74 \\
U-Net & Strong uniform & 0.664 & 0.770 & 5.42 \\
U-Net & Loss-only & 0.704 & 0.805 & 16.13 \\
U-Net & Boundary-only & 0.698 & 0.799 & 13.91 \\
U-Net & \textbf{DASA} & \textbf{0.714} & \textbf{0.813} & 13.09 \\
\midrule
DeepLabV3 & Baseline & 0.633 & 0.742 & 5.56 \\
DeepLabV3 & Strong uniform & 0.716 & 0.813 & 5.55 \\
DeepLabV3 & Loss-only & 0.725 & 0.820 & 13.42 \\
DeepLabV3 & Boundary-only & 0.717 & 0.813 & 14.25 \\
DeepLabV3 & \textbf{DASA} & \textbf{0.740} & \textbf{0.831} & 13.50 \\
\midrule
SegFormer-B0 & Baseline & 0.711 & 0.808 & 8.45 \\
SegFormer-B0 & Strong uniform & 0.686 & 0.787 & 8.18 \\
SegFormer-B0 & Loss-only & 0.730 & 0.824 & 18.40 \\
SegFormer-B0 & Boundary-only & 0.718 & 0.814 & 18.21 \\
SegFormer-B0 & \textbf{DASA} & \textbf{0.730} & \textbf{0.825} & 18.66 \\
\bottomrule
\end{tabular}
\end{table}

\begin{figure}[ht]
    \centering
    \includegraphics[width=0.9\linewidth]{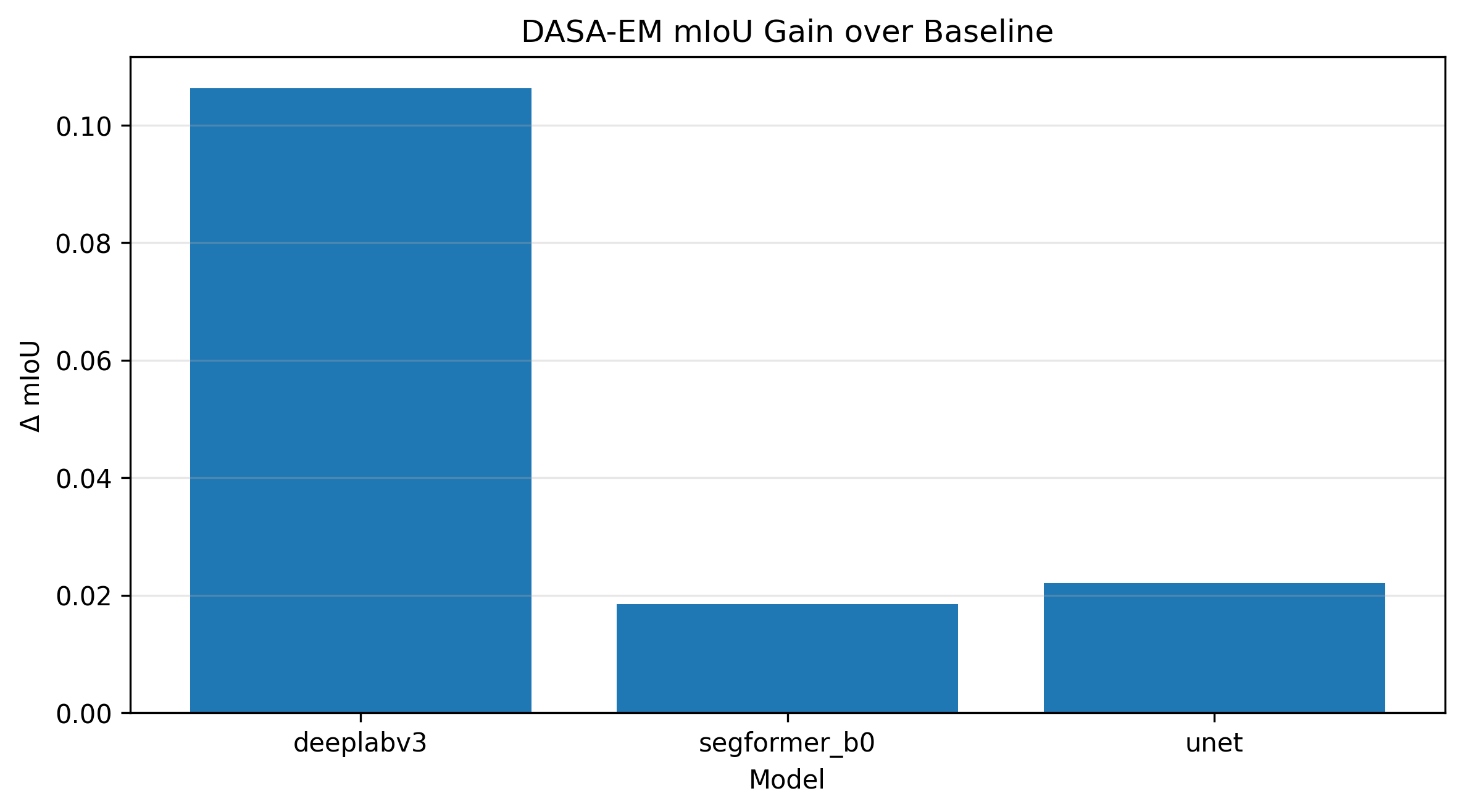}
    \caption{DASA mIoU gain over the non-adaptive baseline on Oxford-IIIT Pet.}
    \label{fig:pet_gain_baseline}
\end{figure}

Figure~\ref{fig:pet_gain_baseline} summarizes the absolute mIoU gain of DASA over the corresponding baseline. The largest improvement occurs for DeepLabV3, but all three architectures benefit from difficulty-aware allocation.

\begin{figure}[ht]
    \centering
    \includegraphics[width=\linewidth]{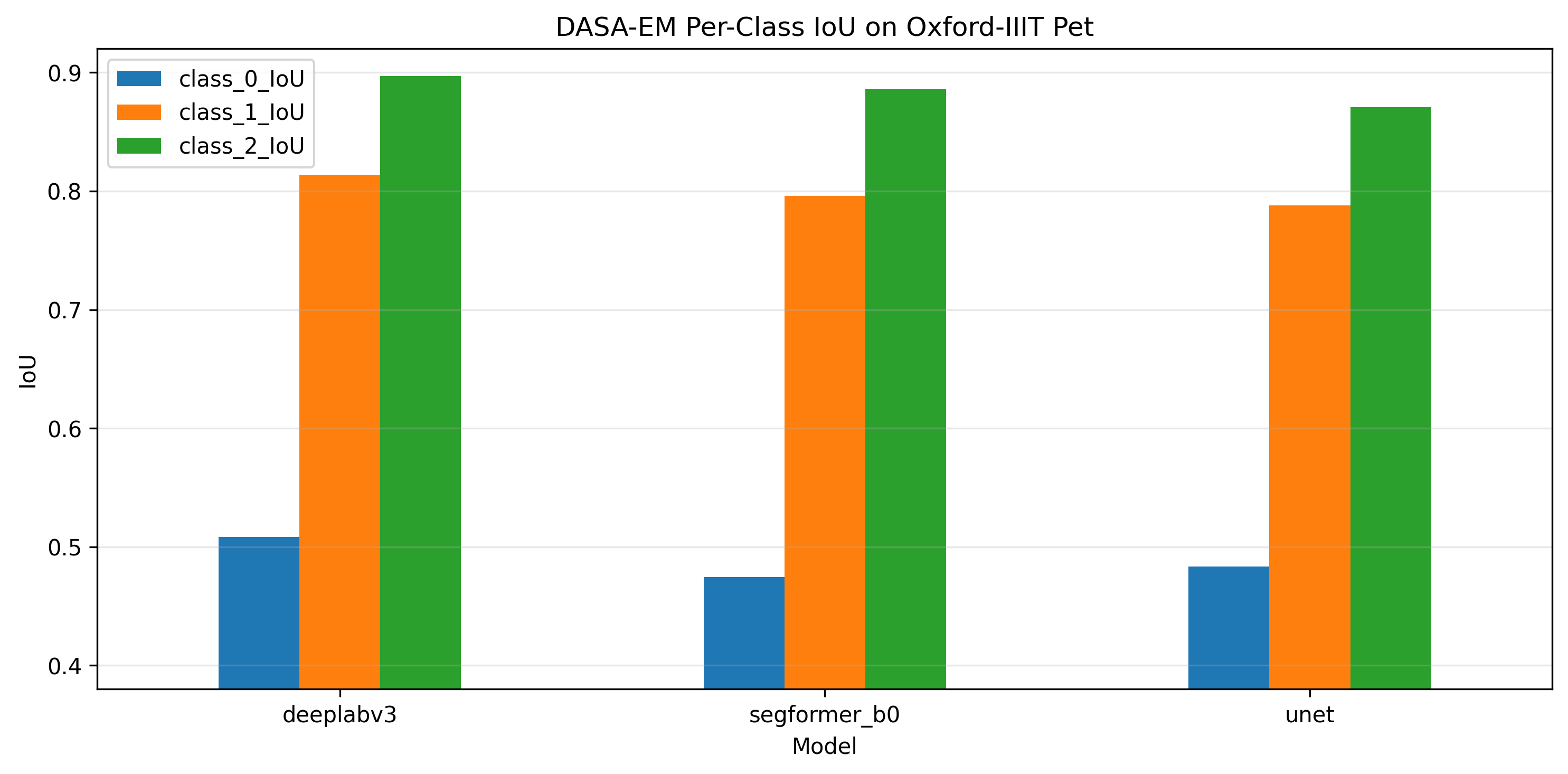}
    \caption{Per-class IoU for DASA on Oxford-IIIT Pet across the three evaluated architectures.}
    \label{fig:pet_per_class_iou}
\end{figure}

Figure~\ref{fig:pet_per_class_iou} shows the class-wise IoU obtained by DASA on Oxford-IIIT Pet. The lower values for class 0 indicate that not all trimap regions have equal difficulty, which supports reporting class-wise behavior in addition to mean scores.

\begin{figure}[ht]
    \centering
    \includegraphics[width=\linewidth]{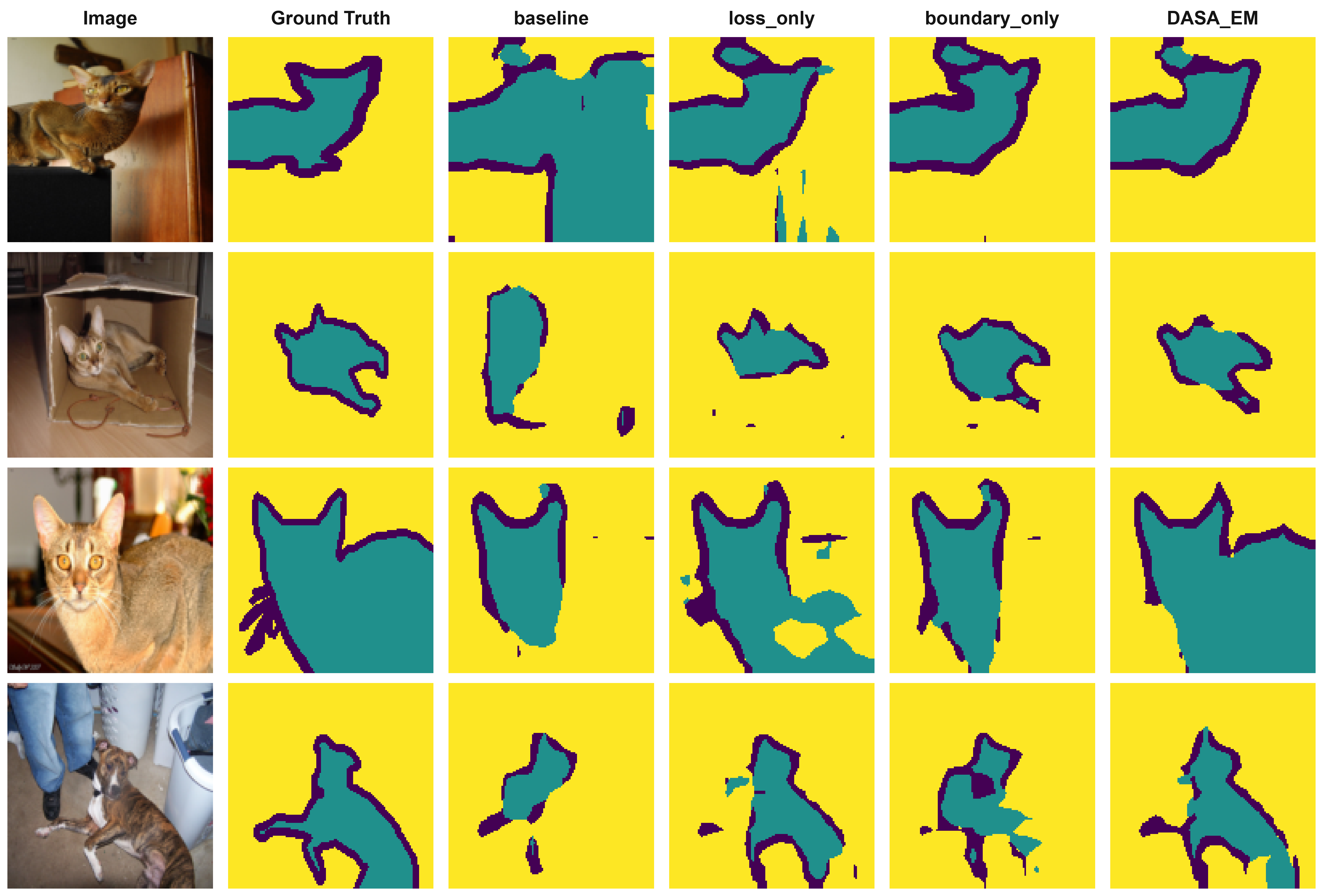}
    \caption{Qualitative Oxford-IIIT Pet segmentation examples for DeepLabV3. The columns compare the input image, ground truth, baseline, loss-only, boundary-only, and DASA predictions.}
    \label{fig:qualitative_pet}
\end{figure}

Figure~\ref{fig:qualitative_pet} provides qualitative DeepLabV3 examples from Oxford-IIIT Pet. The visual comparison complements the aggregate metrics by showing how different allocation strategies affect the predicted trimap masks.

\subsection{Pascal VOC Binary}

Table~\ref{tab:voc_results} shows results on binary Pascal VOC. DASA obtains the highest mIoU for U-Net and SegFormer-B0. For DeepLabV3, DASA is effectively tied with boundary-only allocation in mIoU and achieves the best mDice. Most importantly, DASA produces the highest foreground IoU for all three architectures, indicating that multi-factor allocation is especially helpful for the more difficult non-background class.

This pattern is important because binary foreground-background segmentation can hide errors when background pixels dominate the image. A method may obtain a competitive mean score by segmenting background well while still failing on the object region. DASA improves foreground IoU for all evaluated architectures, which suggests that the allocation rule directs training pressure toward samples that are useful for object recovery rather than merely reinforcing the dominant background class.

\begin{table}[t]
\centering
\caption{Binary Pascal VOC results. FG IoU denotes foreground IoU.}
\label{tab:voc_results}
\begin{tabular}{llcccc}
\toprule
Model & Method & mIoU & mDice & BG IoU & FG IoU \\
\midrule
U-Net & Baseline & 0.491 & 0.586 & 0.763 & 0.219 \\
U-Net & Loss-only & 0.571 & 0.677 & 0.774 & 0.367 \\
U-Net & Rarity-only & 0.600 & 0.709 & 0.778 & 0.423 \\
U-Net & Boundary-only & 0.589 & 0.696 & 0.780 & 0.399 \\
U-Net & \textbf{DASA} & \textbf{0.601} & \textbf{0.713} & 0.760 & \textbf{0.442} \\
\midrule
DeepLabV3 & Baseline & 0.612 & 0.708 & 0.771 & 0.453 \\
DeepLabV3 & Loss-only & 0.634 & 0.725 & \textbf{0.814} & 0.455 \\
DeepLabV3 & Boundary-only & \textbf{0.635} & 0.726 & 0.809 & 0.461 \\
DeepLabV3 & DASA & 0.634 & \textbf{0.731} & 0.787 & \textbf{0.482} \\
\midrule
SegFormer-B0 & Baseline & 0.598 & 0.704 & 0.751 & 0.446 \\
SegFormer-B0 & Rarity-only & 0.621 & 0.716 & \textbf{0.802} & 0.440 \\
SegFormer-B0 & Boundary-only & 0.600 & 0.694 & 0.805 & 0.396 \\
SegFormer-B0 & \textbf{DASA} & \textbf{0.630} & \textbf{0.732} & 0.793 & \textbf{0.468} \\
\bottomrule
\end{tabular}
\end{table}

\subsection{Ablation and Aggregate Trends}

Single-signal adaptive methods often improve over the baseline, confirming that loss, rarity, and boundary information each contain useful difficulty information. However, no single factor dominates across datasets and architectures. Loss-only allocation is a strong competitor for SegFormer-B0 on Oxford-IIIT Pet, rarity-only is strong for U-Net on binary Pascal VOC, and boundary-only is competitive for DeepLabV3 on binary Pascal VOC.

DASA is more stable because it does not depend on one difficulty assumption. Averaged over the three architectures, DASA obtains 0.728 mIoU on Oxford-IIIT Pet and 0.622 mIoU on binary Pascal VOC. It also gives the best foreground IoU on binary Pascal VOC for every architecture, suggesting that multi-factor allocation is particularly useful when the target class is harder than the background.

The ablation results should therefore be read as evidence for complementarity rather than as a claim that every signal is equally important in every setting. Loss can capture current model errors but may over-emphasize noisy or atypical samples. Rarity can protect underrepresented classes but does not know whether the current model already handles a sample well. Boundary complexity captures structural difficulty but can ignore semantic imbalance. The combined score is useful because these failure modes are different.

\subsection{Computational Cost}

DASA introduces overhead because difficulty scores must be estimated and augmentation strengths must be updated. The adaptive methods therefore take longer than baseline training in these experiments. The cost is moderate for the evaluated models and does not require architectural changes. In deployment, the difficulty-estimation frequency can be reduced to trade adaptation quality for runtime.

The runtime results also show that the overhead is not uniform across architectures. Lightweight models can spend a larger proportion of total time on difficulty estimation because the baseline training loop is already fast. Larger models have higher base training cost, so the relative impact of score estimation may be less visible. This means the practical value of DASA should be judged by both accuracy gain and deployment budget. For offline training, the extra cost may be acceptable when foreground recovery is important. For constrained settings, updating difficulty scores less frequently is a natural compromise.

\section{Discussion}

The experiments support three observations. First, adaptive augmentation is usually preferable to a fixed uniform policy, especially when the dataset contains heterogeneous sample difficulty. Second, the best single difficulty signal varies by dataset and architecture, which motivates a combined estimator. Third, DASA's strongest and most consistent advantage appears on difficult foreground regions. This is important for segmentation tasks where the background dominates pixel counts but foreground classes determine practical usefulness.

The results do not imply that every difficult sample should always receive severe transformations. Excessive augmentation can distort labels, destroy boundary alignment, or create unrealistic training examples. DASA is designed to avoid this by mapping difficulty into bounded strengths rather than allowing unlimited transformations. This bounded interpretation is important for segmentation, where geometric changes must preserve pixel-level correspondence between images and masks.

The current implementation uses fixed aggregation weights. This makes the method simple and transparent, but it may not be optimal for all datasets. A natural extension is to learn the weights or update them based on validation performance. Another limitation is that ambiguity estimation requires additional stochastic forward passes. Future work should examine cheaper uncertainty indicators and larger-scale multi-class benchmarks.

\section{Conclusion}

This paper introduced Difficulty-Aware Sample Allocation, a multi-factor adaptive augmentation framework for semantic segmentation. DASA combines prediction ambiguity, training loss, class rarity, and boundary complexity to allocate stronger augmentation to more difficult samples. Experiments with U-Net, DeepLabV3, and SegFormer-B0 on Oxford-IIIT Pet and binary Pascal VOC show that DASA improves over standard training, competes strongly with single-signal adaptive baselines, and consistently improves foreground segmentation on Pascal VOC. Because DASA is architecture-agnostic and requires no network modifications, it can be integrated into existing segmentation pipelines with limited implementation cost.

\subsection*{Funding} 

The authors received no funding for this study.

\subsection*{Ethics, Consent to Participate, and Consent to Publish declarations} 

Not applicable.

\bibliography{references}

\end{document}